\documentclass[letterpaper,10pt,conference]{ieeeconf}
\IEEEoverridecommandlockouts
\pdfoutput=1

\usepackage[noadjust]{cite}  %
\usepackage{xparse} 
\usepackage{amsmath} 
\usepackage{amssymb}
\usepackage{amsfonts} 
\usepackage{bbm}
\usepackage{color}
\usepackage{mathtools}
\usepackage{multirow}
\usepackage{graphicx}
\usepackage{subcaption}
\usepackage{wrapfig}
\usepackage{calc}
\usepackage{balance}
\usepackage{booktabs}
\usepackage{url}
\usepackage{algpseudocode}
\usepackage{algorithm}
\usepackage{tabularx}
\usepackage{flushend}  %

\makeatletter
\let\NAT@parse\undefined
\makeatother
\usepackage[colorlinks]{hyperref}
\usepackage[capitalize]{cleveref}

\usepackage{xparse}

\NewDocumentCommand\bbm{}{ \begin{bmatrix} }
\NewDocumentCommand\ebm{}{ \end{bmatrix} }

\NewDocumentCommand\Matrix{m}{ \boldsymbol{\mathbf{#1}} }

\NewDocumentCommand\LieGroupSE{m}{ \mathrm{SE}(#1) }

\NewDocumentCommand\CoordinateFrame{m}{ \underrightarrow{\Matrix{\mathcal{F}}}_{#1} }

\NewDocumentCommand\DE{}{\mathcal{D}_E}

\NewDocumentCommand\pith{}{\pi_\theta}
\NewDocumentCommand\piE{}{\pi_E}

\title{\LARGE\bf Seeing All the Angles: Learning Multiview Manipulation\\ Policies for Contact-Rich Tasks from Demonstrations}
\author{Trevor Ablett, Yifan Zhai, and Jonathan Kelly$^\dagger$
\thanks{All authors are with the Space \& Terrestrial Autonomous Robotic Systems (STARS) Laboratory at the University of Toronto Institute for Aerospace Studies (UTIAS), Toronto, Canada {\tt <firstname>.<lastname>@robotics.utias.utoronto.ca}.}
\thanks{$^\dagger$Jonathan Kelly is a Vector Institute Faculty Affiliate. This research was supported in part by the Canada Research Chairs program.}}

\begin{document}
\maketitle
\thispagestyle{empty}
\pagestyle{empty}

\begin{abstract}
	Learned visuomotor policies have shown considerable success as an alternative to traditional, hand-crafted frameworks for robotic manipulation. 
	Surprisingly, an extension of these methods to the multiview domain is relatively unexplored. 
	A successful multiview policy could be deployed on a mobile manipulation platform, allowing the robot to complete a task regardless of its view of the scene.
	In this work, we demonstrate that a multiview policy can be found through imitation learning by collecting data from a variety of viewpoints.
	We illustrate the general applicability of the method by learning to complete several challenging multi-stage and contact-rich tasks, from numerous viewpoints, both in a simulated environment and on a real mobile manipulation platform. 
	Furthermore, we analyze our policies to determine the benefits of learning from multiview data compared to learning with data collected from a fixed perspective. 
	We show that learning from multiview data results in little, if any, penalty to performance for a fixed-view task compared to learning with an equivalent amount of fixed-view data.
	Finally, we examine the visual features learned by the multiview and fixed-view policies. 
	Our results indicate that multiview policies implicitly learn to identify spatially correlated features.
\end{abstract}

\section{Introduction}
\label{sec:introduction}

The use of end-to-end visuomotor policies, in which observations are mapped directly to actions through a learned model, has emerged as an effective alternative to the traditional sense-plan-act approach for many robotic domains including autonomous driving \cite{pomerleauALVINNAutonomousLand1989,bojarskiEndEndLearning2016} and manipulation \cite{levineEndtoendTrainingDeep2016,zhangDeepImitationLearning2018, laskeyDARTNoiseInjection2017}. 
Visuomotor policies are particularly appealing for manipulation because programming a robot to complete even relatively basic tasks can pose a major challenge.
Most research on learning end-to-end manipulation policies, where the inputs to the policy are easily-acquired camera images and proprioceptive sensor data, has focused on a fixed-base arm and a fixed camera viewpoint.
If these policies are na\"{i}vely rolled out for a task that requires the camera angle or base position to change even slightly, as is often the case for a mobile manipulator, one would not expect the policies to succeed. %
We seek to learn highly generalizable policies that are not brittle in the face of perturbations to the viewpoint and base position.

\begin{figure}
	\centering
	\vspace{2mm}
	\setlength{\fboxsep}{0pt}%
	\setlength{\fboxrule}{1pt}%
	\includegraphics[width=\columnwidth]{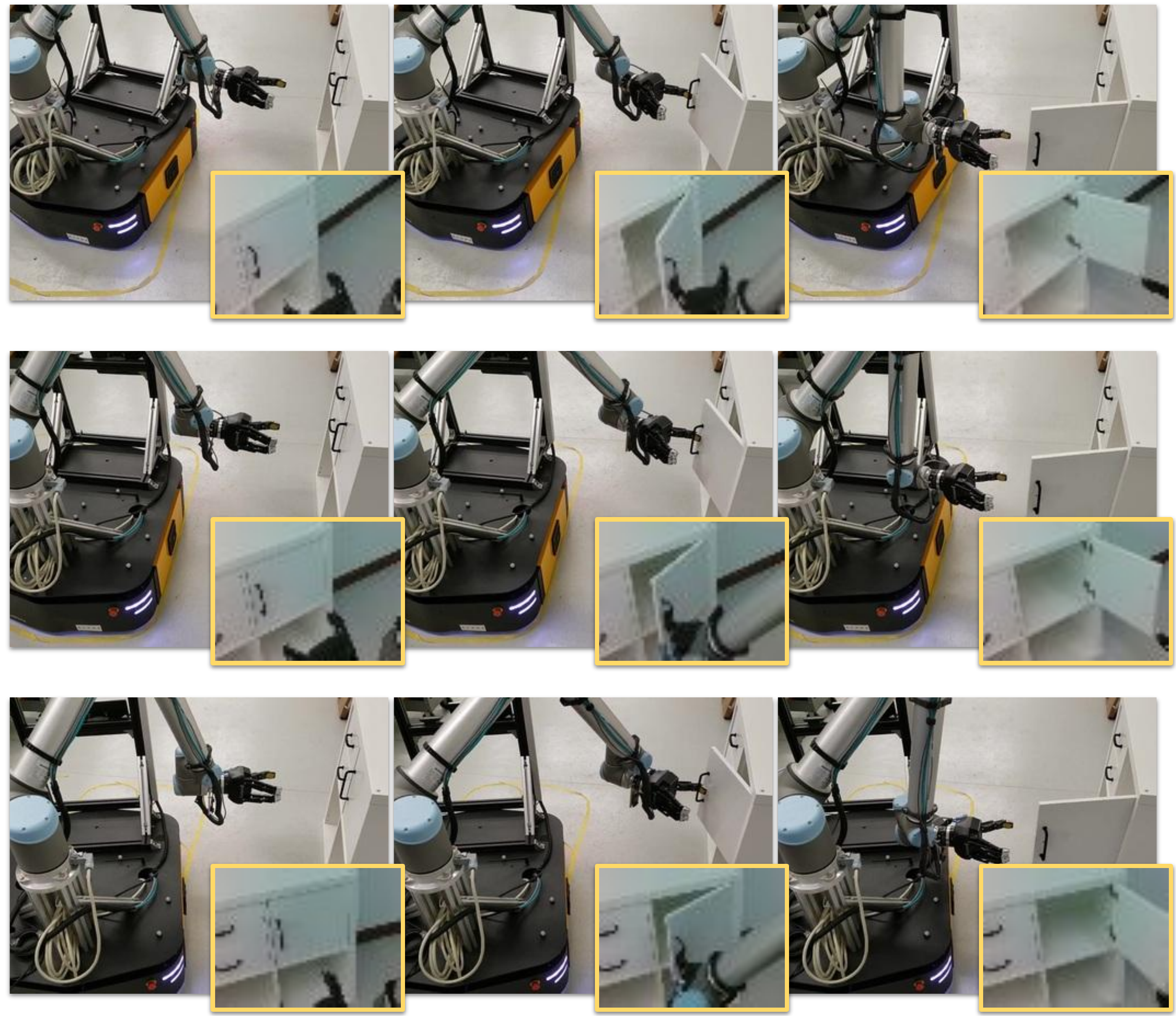}
	\caption{Snapshots of three successful trajectories involving different viewpoints for our real-world cabinet-opening task, executed by a single policy. Each trajectory is shown left-to-right.
	The yellow boxes highlight individual $64 \times 48$ RGB input image frames. Our end-to-end policy is able to generalize to the different images and base poses corresponding to each viewpoint.}
	\label{fig:real_both_views_runs}
	\vspace{-3mm}
\end{figure}

In this work, we investigate the application of supervised imitation learning for generating end-to-end \textit{multiview} policies for complex, contact-rich tasks.
Specifically, we create datasets containing trajectories with varying base poses, allowing single policies to learn to complete tasks from a variety of viewpoints (see \cref{fig:real_both_views_runs}).
Policies learned in this way can be directly applied to mobile manipulators in conjunction with a separate navigation policy that moves the mobile base to the vicinity of the manipulation task workspace \cite{iriondoPickPlaceOperations2019}. 
Our main contributions are to answer the following questions:
\begin{enumerate}
	\item How does supervised imitation learning perform in a series of challenging contact-rich tasks in the multiview domain?
	\item \label{enum:mult_vs_fix} Will a policy trained with multiview data be penalized when performing an equivalent fixed-perspective task, compared to a policy trained with an equal amount of exclusively fixed-perspective data?
	\item How far can fixed-base and multiview policies be pushed beyond their training distributions?
	\item Compared with a fixed-view policy, do the features learned by a multiview policy show greater spatial correlation between different views?
\end{enumerate}

\begin{figure*}[t!]
	\centering
	\smallskip
	\setlength{\fboxsep}{0pt}%
	\setlength{\fboxrule}{1pt}%
	\includegraphics[width=.88\textwidth]{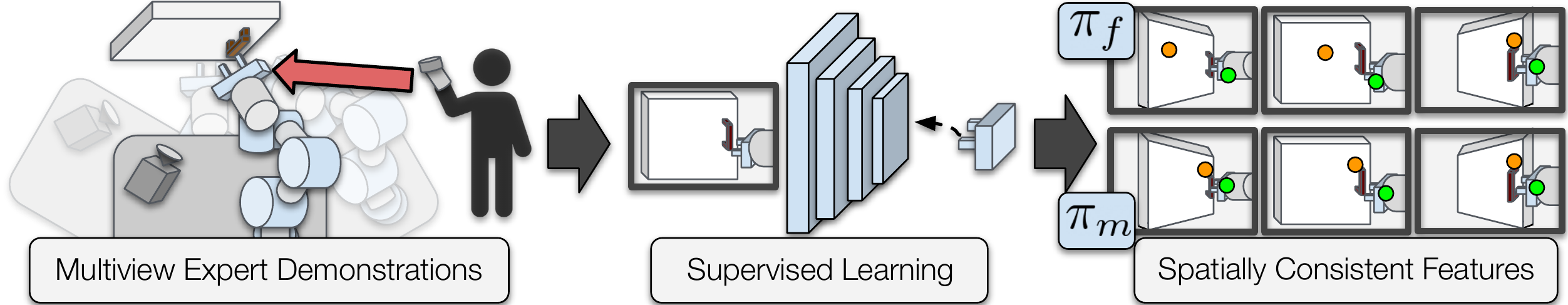}
	\vspace{1mm}
	\caption{Our system for generating multiview policies: after collecting expert demonstrations from several viewpoints, we train a deep neural network policy that generalizes to multiple views. The learned visual features consistently appear on the same parts of objects.}
	\label{fig:system_outline}
	\vspace{-3mm}
\end{figure*}

Somewhat surprisingly, we demonstrate that multiview policies have near-equal performance to fixed-view policies in fixed-view domains, motivating their use in any case where a mobile base is present.
We do not explicitly encode view-invariance in the loss function or policy architecture, and instead show that end-to-end multiview policies can be effectively trained implicitly by modifying the dataset.

\section{Related Work}

In this section, we begin by examining existing work on end-to-end policy learning for robotics, followed by learning-based mobile manipulation and other multiview manipulation research.
We close with a brief discussion of supervised imitation learning, also known as behaviour cloning.

The use of deep visuomotor policies that map raw observations to actions has exploded in popularity recently, largely owing to representational power and generalization capabilities of deep convolutional neural networks \cite{levineEndtoendTrainingDeep2016, finnGuidedCostLearning2016, codevillaEndtoEndDrivingConditional2018}. 
Encoding a policy this way offers the advantage of being able to learn directly from data, given either expert actions or an external reward signal, without requiring accurate world state information and a handcrafted behaviour policy.
Such policies have the downside of being limited to operating on data that closely resemble the data they were trained on. 
In this work, we expand the training dataset to include multiple viewpoints, substantially improving the robustness of the learned policies.

Our approach can be compared to other learning-based methods used for mobile manipulation.
The system developed in \cite{bajracharyaMobileManipulationSystem2020} applies several learning techniques, combined with connected motion primitives, to complete several tasks. 
View-invariance is encoded within separate object recognition and planning modules. Our method, in comparison, uses an end-to-end approach for completing tasks given raw sensor data only.

Several groups have attempted to learn policies that control both a mobile base and a manipulator simultaneously \cite{wangLearningMobileManipulation2020, welscheholdLearningMobileManipulation2017, kindleWholeBodyControlMobile2020}.
In each of these cases, the authors make assumptions about the availability of lower-level state information \cite{welscheholdLearningMobileManipulation2017, wangLearningMobileManipulation2020}, or confine their systems to perform only relatively simple reaching tasks \cite{wangLearningMobileManipulation2020, kindleWholeBodyControlMobile2020}.
While we do not attempt to control the base during task execution, we generate policies that can complete challenging, contact-rich tasks \textit{without} access to privileged state information.
The authors of \cite{laskeyLearningRobustBed2017} learn a mobile manipulation bed-making task with imitation learning, but fiduciary markers are required at the base positions to ensure precise localization.

There has been some interest in learning policies that generalize to multiple views even when a fixed-base manipulator is employed.
Through view synthesis, simulated views \cite{aminiLearningRobustControl2020} or their latent representations \cite{eslamiNeuralSceneRepresentation2018} can be used for generating higher-quality policies.
This approach has two major drawbacks: it requires a potentially prohibitive amount of training data and it operates on the assumption that all parts of images are relevant.
Our method learns policies that output control signals given raw images, ensuring that only the parts of the scene relevant for control are extracted.

Other research has investigated the use of multiview representations learned with contrastive losses, determined from either time-aligned sequences from multiple camera views \cite{sermanetTimeContrastiveNetworksSelfSupervised2018, dwibediLearningActionableRepresentations2018, maedaVisualTaskProgress2020} or via pre-existing object-recognition software \cite{florenceDenseObjectNets2018, florenceSelfSupervisedCorrespondenceVisuomotor2020}.
Our method does not assume access to any extra information beyond raw RGB-D sensor data\footnote{We use depth information because it is easily acquired along with images using off-the-shelf RGB-D sensors.}, but presumably, these representations could be used to improve the learned policies in our work.

In \cite{sadeghiSim2RealViewpointInvariant2018}, the authors applied domain randomization \cite{tobinDomainRandomizationTransferring2017} to learn policies that are able to complete a real-world multiview reaching task.
The final policy is able to generalize to inputs from novel viewpoints.
In contrast, our tasks require significantly higher dexterity than reaching alone.

Behaviour cloning (BC) is the common name given to imitation learning treated as supervised learning \cite{bainFrameworkBehaviouralCloning1996, pomerleauALVINNAutonomousLand1989}: after collecting an expert dataset, a policy is trained to regress to expert actions given the corresponding observations. 
A core assumption of supervised learning is that the training and test data are independently and identically distributed (IID). In BC, this translates to assuming that the policy dataset, generated by running the policy, is drawn from the same distribution as as the expert dataset. 
Unfortunately, in general this assumption is violated \cite{rossReductionImitationLearning2011}, but the problem can be mitigated by manually increasing the coverage of the expert dataset \cite{pomerleauALVINNAutonomousLand1989,zhangDeepImitationLearning2018, laskeyDARTNoiseInjection2017} or by employing an intervention-based strategy \cite{ablettFightingFailuresFIRE2020}. 
We investigate the effects of including and excluding multiple views and base poses in the expert dataset for both multiview and fixed-base tasks.

\section{Problem Formulation}

We formulate our problem as a Markov Decision Process (MDP). Our goal is to learn a deterministic policy $\pi_\theta: O \rightarrow A$, parameterized by $\theta$, for environment observations $o \in O$ and actions $a \in A$. 
Instead of maximizing a reward, in imitation learning, we attempt to match a learned policy $\pith$ to an expert policy $\piE$.
In our case, we do not assume we have direct access to $\piE$ and instead only have samples of human-generated demonstrations $\DE := \{\tau_1, \dots, \tau_n, \dots, \tau_N\}, \tau_n := \{(o_0, a_0), \dots, (o_t, a_t), \dots, (o_{T-1}, a_{T-1}) \}$, where $T$ is the task horizon length. 
The initial observation $o_0$ is sampled from a pre-defined distribution $p(o_0)$.

We train our policies using behaviour cloning.
The policy $\pith$ can be trained by minimizing the mean squared error,
\begin{align} \label{eq:bc_mse}
	\min_{\theta} \sum_{(o,a) \in \DE} \left(\pi_\theta(o) - a \right)^2.
\end{align}

In our work, each individual task or environment $\mathcal{T}$ is a separate MDP that can be considered to be either \textit{multiview} ($\mathcal{T}_m$) or \textit{fixed-view} ($\mathcal{T}_f$), with $\pi_m$ denoting a policy trained with data from $\mathcal{T}_m$ (and $\pi_f$ for $\mathcal{T}_f$), where we omit $\theta$ for convenience. 
For $\mathcal{T}_f$, we can define the observation generating process $g_{O,f} : S_\mathcal{M} \times S_i \rightarrow O_{\mathcal{T}_f}$, noting that $o_{\mathcal{T}_f} \in O_{\mathcal{T}_f}$ are generated by an unknown function $g_{O,f}$ of the underlying states of our manipulator $s_\mathcal{M} \in S_\mathcal{M}$ and task-relevant objects $s_i \in S_i$.
The initial states of each episode, $s_{\mathcal{M}, 0}$ and $s_{i,0}$, are uniformly randomized within predefined constraints. %
In contrast, for the multiview case, we define $g_{O,m} : S_\mathcal{M} \times S_i \times S_b \rightarrow O_{\mathcal{T}_m}$, where we have added the state of the base of our robot $s_b \in S_b$, noting that $s_b$ is randomized only \textit{between} episodes.
In our formulation, measurements $o \in O$ are acquired from a sensor attached to the robot base and from the arm itself, so changing $s_{b,0}$ affects $s_{\mathcal{M},0}$, the view of task relevant objects, and the set of actions that are able to `solve' the task.

\section{Multiview Training and Shared Information} 
\label{sec:multiview_data_analysis}

As stated in Section \ref{sec:introduction}, we are interested in the comparison between a fixed-base task $\mathcal{T}_f$ and an equivalent multiview version $\mathcal{T}_m$, as well as policies $\pi_f$ and $\pi_m$ trained on observations $O_{\mathcal{T}_f}$ and $O_{\mathcal{T}_m}$. 
It is important to note that because $S_\mathcal{M}$ and $S_i$ are shared between these environments and $\text{dim}(O_{\mathcal{T}_m}) = \text{dim}(O_{\mathcal{T}_f})$, we can generate actions from $\pi_m$ or $\pi_f$ with both $O_{\mathcal{T}_m}$ and $O_{\mathcal{T}_f}$.

\subsection{Comparing $\mathcal{T}_m$ and $\mathcal{T}_f$}

Considering the sizes of the sets of possible states for $\mathcal{T}_m$ and $\mathcal{T}_f$ leads to a well-known challenge in prediction problems, the curse of dimensionality \cite{bellmanDynamicProgramming1957}: since $\text{dim}(S_\mathcal{M} \times S_i \times S_b) > \text{dim}(S_\mathcal{M} \times S_i)$, we should require more training examples from $O_{\mathcal{T}_m}$ to learn $\pi_m$ than from $O_{\mathcal{T}_f}$ to learn $\pi_f$ (i.e., to achieve the same success rate). Said differently, more examples are required to adequately `cover the space' $O_{\mathcal{T}_m}$ than to cover $O_{\mathcal{T}_f}$.

A natural conclusion is that, given the same quantity of training data, $\pi_m$ will perform \textit{worse} than $\pi_f$ on $\mathcal{T}_f$, for two separate but related reasons: i) $\pi_m$ is required to learn a higher-dimensional problem than $\pi_f$, and ii) $\pi_m$ is provided with less (or possibly no) $o_{\mathcal{T}_f} \in O_{\mathcal{T}_f}$ at training time.

\subsection{When Multiview Data Helps} 
\label{sec:correlation_views}

Implicit in the above conclusion is the assumption that, for a specific task, the distributions of expert actions $p(a_E\,|\, s_b = \alpha)$ and $p(a_E\,|\, s_b = \beta)$ for two different base poses, $\alpha, \beta \in S_b$, are independent.
However, this is not true for our problem, or for many other supervised learning tasks. The distributions of observations (and actions) for poses that are `nearby' in the state space, $S_b$, must have nonzero mutual information, as well as smooth state, observation, and action spaces.
If this were not the case, multiview policies would be unable to generalize to new poses.
We experimentally show that our policies are able to generalize to new poses (in \cref{sec:perf_experiments}) and even to out-of-distribution data (in \cref{sec:ood_experiments}).
We consider the task-dependent mutual information $I(A_{E, \alpha}; A_{E, \beta})$ between $A_{E, \alpha} \sim p(a_E\,|\, s_b = \alpha)$ and $A_{E, \beta} \sim p(a_E\,|\, s_b = \beta)$ below.

For tasks where $I(A_{E, \alpha}; A_{E, \beta})$ is large in general, $\pi_m$ will provide little benefit over $\pi_f$ in $\mathcal{T}_m$.
However, when $I(A_{E, \alpha}; A_{E, \beta})$ is small in general, $\pi_m$ may be prohibitively costly to learn and suffer compared with $\pi_f$ in $\mathcal{T}_f$. This issue arises due to the increased number of expert demonstrations needed to cover the space of $O_{\mathcal{T}_m}$.
For this reason, we expect $\pi_m$ will provide the most benefit, compared to $\pi_f$ learned from an equivalent amount of data,  when $I(A_{E, \alpha}; A_{E, \beta})$ falls somewhere in the middle---each base pose generates similar observations and requires a similar, but not identical, trajectory of actions to allow successful completion of the task (see \cref{fig:real_both_views_runs}). 
As an example, consider a lifting task, where the robot has to lift an object sitting on a table. 
If, when varying $s_b$, the object poses in the set $S_{i,0}$ remain the same relative to both the imaging sensor and the robot base, a multiview policy would provide little benefit.
Conversely, consider a door opening task, where the robot has to open a cabinet door: the initial pose of the cabinet $s_{i,0}$ does not change relative to the world.
Therefore, if $s_b$ is changed, the initial pose of the cabinet, relative to the imaging sensor and the robot base, will necessarily change, and a fixed-view policy will likely fail.

We can rephrase our first two experimental questions (see \cref{sec:introduction}): provided that a task has an upper bound on the range of base poses to consider, is there sufficient mutual information between trajectories at different (but `nearby') poses to enable a policy $\pi_m$ to be learned that not only performs adequately in $\mathcal{T}_m$, but performs comparably to $\pi_f$ in $\mathcal{T}_f$, given the same amount of training data? 
That is, if we reduce the sampling density (of expert demonstrations), is the mutual information between those demonstrations sufficient to ensure that $\pi_m$ performs well in $\mathcal{T}_m$ \emph{and} similarly to $\pi_f$ in $\mathcal{T}_f$? 
We explore this question in \cref{sec:perf_experiments}.

Finally, despite nonzero mutual information, it remains true that $\pi_m$ must learn in the larger space of $O_{\mathcal{T}_m}$, compared to $\pi_f$ and  $O_{\mathcal{T}_{f}}$.
If $\pi_m$ relies on mutual information, we would expect that many of the visual features learned by $\pi_m$ would consistently refer to the same parts of the scene, regardless of viewpoint, a possibility which we examine in \cref{sec:feature_analysis}.

\section{Methodology} 
\label{sec:data_collection}

\begin{figure}[t!]
	\centering
	\smallskip
	\setlength{\fboxsep}{0pt}%
	\setlength{\fboxrule}{1pt}%
	
	\begin{subfigure}{0.49\columnwidth}
		\includegraphics[width=\textwidth - 2pt]{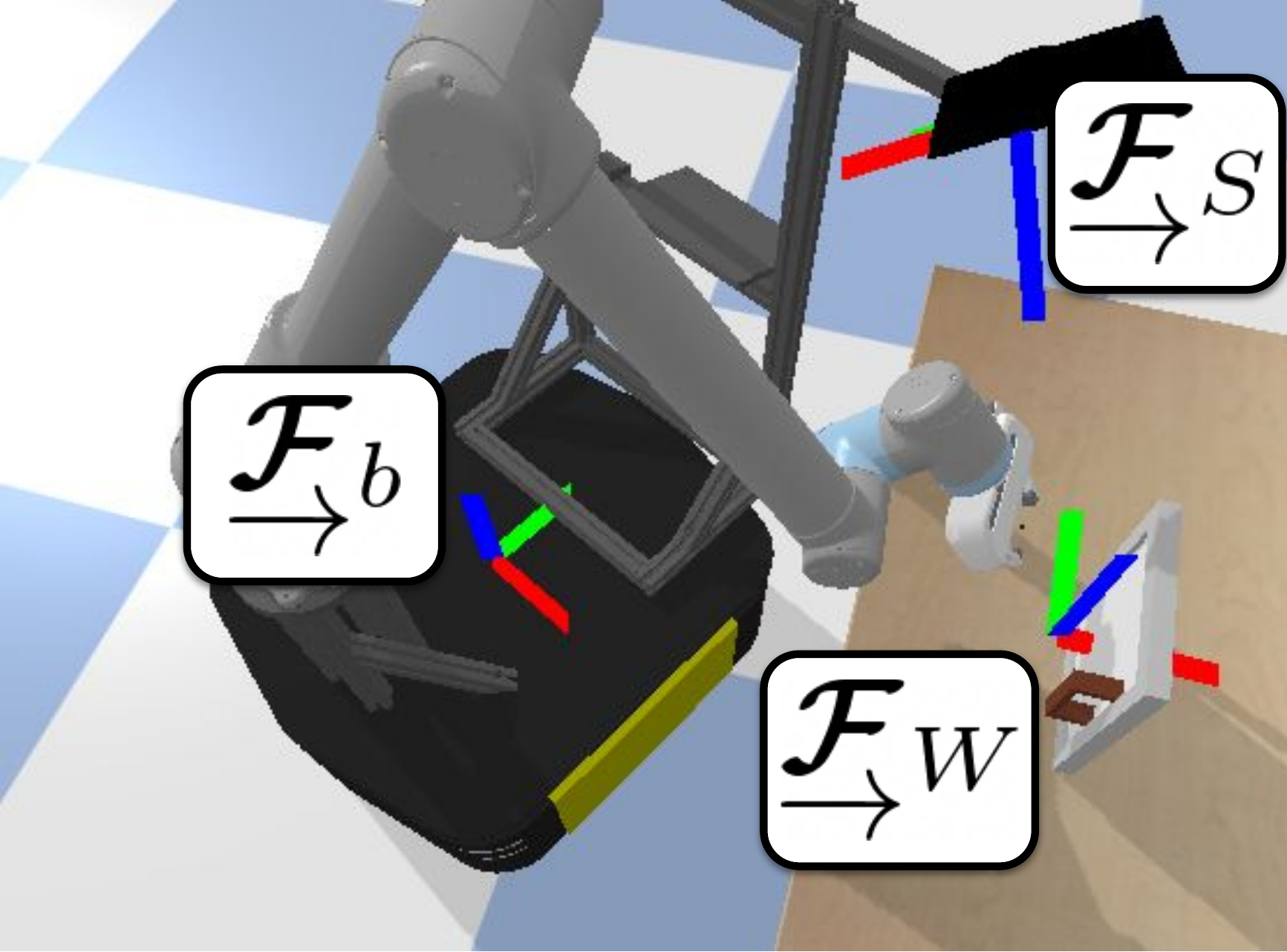}
	\end{subfigure}
	\begin{subfigure}{0.49\columnwidth}
		\includegraphics[width=\textwidth - 2pt]{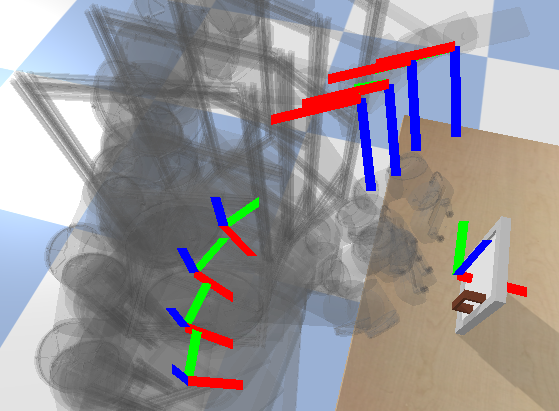}
	\end{subfigure}
	
	\vspace{1mm}
	\begin{subfigure}{.995\columnwidth}
		\includegraphics[width=\textwidth - 2pt]{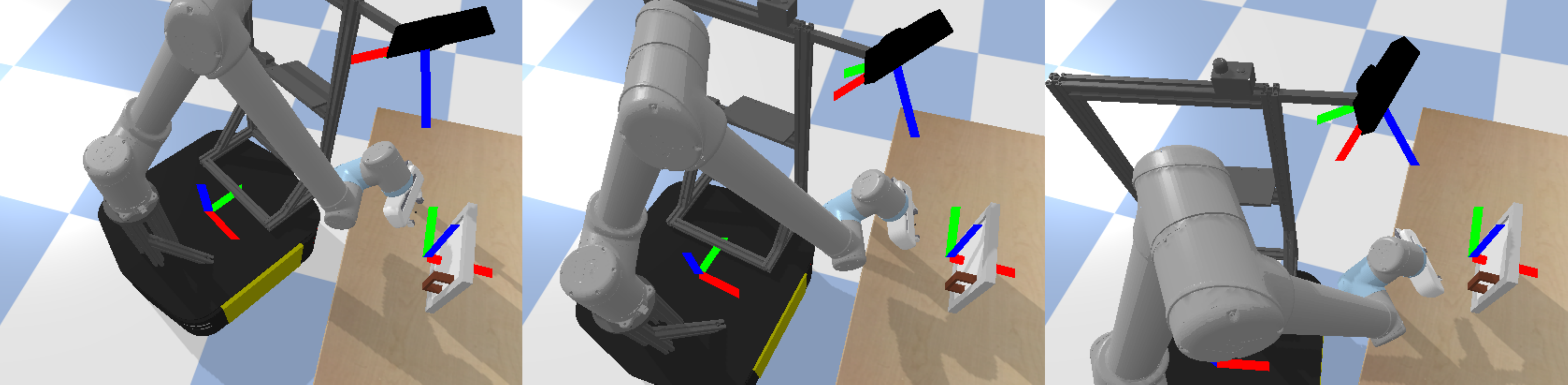}
	\end{subfigure}
	
	\caption{Top left: Relevant robot and task reference frames. Top right: A set of four base poses, shown to illustrate the semi-circle of possible poses. Bottom: Base poses where $b_{\phi}$ is set to $b_{\phi,\text{min}}$, $0.5 (b_{\phi,\text{min}} + b_{\phi,\text{max}})$, and  $b_{\phi,\text{max}}$.}
	\label{fig:base_pose_frames_all}
	\vspace{-5mm}
\end{figure}
\label{sec:auto_gen_new_views}

As previously noted, we assume that we have access to an ``approach'' policy that is capable of moving the mobile base to a pose where the task-relevant objects are i) in view of the base-mounted sensor and ii) within the reachable workspace of the manipulator.

We use an automated process to generate randomized base poses for each new training episode. 
We require i) a rough estimate of the transform from the robot base frame to the sensor (often provided, and easily acquired through off-the-shelf calibration), ii) a rough estimate of the robot base pose in the fixed world reference frame $\CoordinateFrame{W}$ (the estimate from wheel odometry is adequate), iii) a pre-selected centre point in $\CoordinateFrame{W}$, and iv) the desired distance (again, approximate) between the camera frame origin and the centre point in $\CoordinateFrame{W}$. 
We desire poses of the mobile base where the main optical axis of the camera sensor, at $\CoordinateFrame{S}$, always very nearly intersects with the selected centre point in $\CoordinateFrame{W}$. 
Each pose in the feasible set lies on a circle, as shown in \cref{fig:base_pose_frames_all}. 
Since the base poses are in $\LieGroupSE{2}$, they can be defined by $b_{\phi}, b_x$ and $b_y$: new poses are generated by randomly sampling $b_{\phi} \sim U(b_{\phi, \text{min}}, b_{\phi, \text{max}})$ (where $b_{\phi, \text{min}}$ and $b_{\phi, \text{max}}$ are set to ensure that there are no collisions with the environment) and computing the appropriate corresponding $b_x$ and $b_y$ using the constraints outlined above.
After sampling and solving for the full base pose, we add a small amount of uniform random noise to the pose values (ensuring that task-relevant objects remain in view), with the aim of increasing the robustness of our learned policy. 
Importantly, our learned policies do not require direct knowledge of the base-to-camera transform or the workspace-to-base transform---this information is used only during the autonomous view generation process.

Our method also requires the collection of a dataset $\DE$ of expert trajectories of observation-action pairs $(o, a)$ acquired through teleoperation. 
We only require that the demonstrations are collected without an operator in view of the imaging sensor, though we note that this constraint exists for all visuomotor imitation learning methods.

\section{Experimental Setup} \label{sec:experimental_setup}
\begin{figure}
	\centering
	\smallskip
	\setlength{\fboxsep}{0pt}%
	\setlength{\fboxrule}{1pt}%
	\includegraphics[width=\columnwidth]{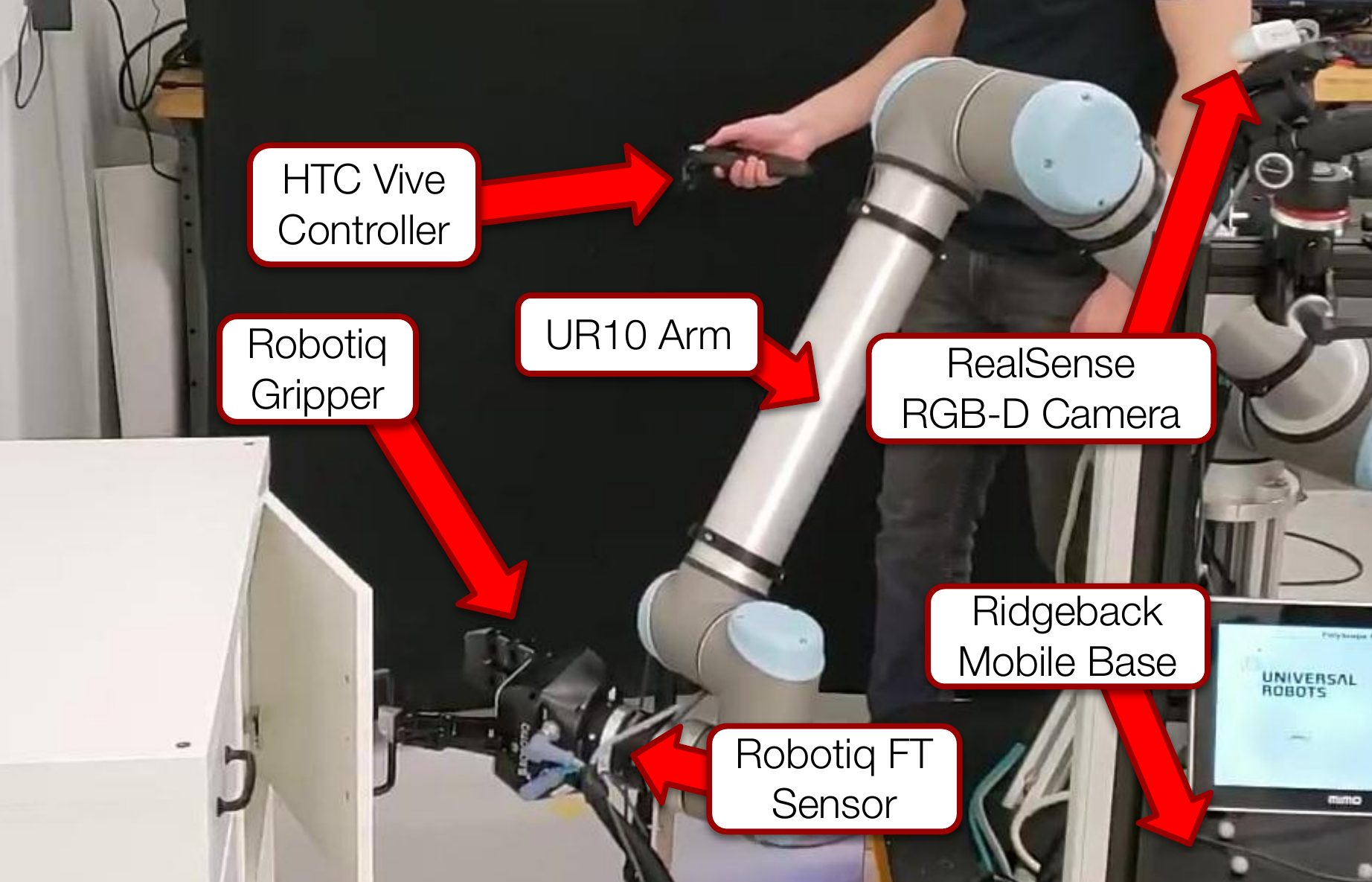}
	\caption{Our experimental setup in the real world. Pictured is our mobile manipulation platform as described in \cref{sec:exp_hardware}, as well as a human expert in the process of collecting a demonstration for our \textit{DoorReal} task. See attached video for example demonstrations.}
	\label{fig:real_experimental_setup}
	\vspace{-5mm}
\end{figure}
\begin{table*}
	\renewcommand{\arraystretch}{1.05}
	\centering
	\caption{Environments considered in this work. Demo time is the cumulative real time of the 200 demonstrations in the multiview version of each expert dataset and the $p(o_0)$ params are the initial conditions of the environment that are randomized between episodes. The ``base $b_{\phi}$ range" corresponds to $b_{\phi, \text{max}} - b_{\phi, \text{min}}$, as described in \cref{sec:auto_gen_new_views}.}
	\begin{tabularx}{\linewidth}{l|XlXXl}
		Environment & Objective & Demo time & $p(o_0)$ params & $p(o_0)$ (ranges) & Actions
		\\\midrule
		\textit{LiftSim}           & Reach block and lift above 7.5cm                & 18m23s & base pose, block pose                        & base $b_{\phi}$ range: $45^\circ$, block center in 25cm$\times$25cm box       & Trans vel, grip \\
		\textit{StackSim}          & Stack blue block on green block                 & 25m05s & base pose, blue block pose, green block pose & base $b_{\phi}$ range: $45^\circ$, block centers in 15cm$\times$15cm box      & 6-DOF vel, grip \\
		\textit{PickAndInsertSim}  & Grasp cylinder and insert in hole ($<$1mm tol.) & 14m23s & base pose, cylinder pose                     & base $b_{\phi}$ range: $45^\circ$, cylinder center in 2.5cm$\times$2.5cm box  & 6-DOF vel, grip \\
		\textit{DoorSim}           & Grasp door handle, open $>$90$^\circ$           & 25m17s & base pose, initial gripper pose              & base $b_{\phi}$ range: $45^\circ$, gripper: in 12cm$\times$5cm$\times$5cm box & 6-DOF vel, grip \\
		\textit{PickAndInsertReal} & Grasp cylinder and insert in hole ($<$1mm tol.) & 28m30s & base pose, cylinder pose                     & base $b_{\phi}$ range: $35^\circ$, cylinder center in 2.5cm$\times$2.5cm box  & 6-DOF vel, grip  \\
		\textit{DoorReal}          & Hook door handle, open $>$90$^\circ$            & 30m56s & base pose, initial gripper pose                     & base $b_{\phi}$ range: $35^\circ$, gripper: in 12cm$\times$5cm$\times$5cm box  & 6-DOF vel \\
		\textit{DrawerReal}        & Hook drawer handle, open within 2cm of max      & 32m23s & base pose, initial gripper pose                     & base $b_{\phi}$ range: $35^\circ$, gripper: in 12cm$\times$5cm$\times$5cm box & 6-DOF vel \\
	\end{tabularx}
	\label{tab:env_details}
	\vspace{-4mm}
\end{table*}

In this section, we describe our experimental design, including the hardware used, the parameters of our tasks, and how we train our policies.

\subsection{Hardware} \label{sec:exp_hardware}

We carry out experiments on both simulated and real versions of our mobile manipulation platform, shown in \cref{fig:real_experimental_setup}.
Our real platform has a Robotiq 3-finger gripper, while our simulated platform in PyBullet \cite{coumans2019} uses either a PR2 gripper or a Franka Emika Panda gripper due to the difficulty of simulating the 3-finger gripper.

On our physical platform, we employ a simple compliant controller using a Robotiq FT-300 force-torque sensor, allowing our policies to operate safely in our contact-rich tasks.
Our robot is controlled using off-the-shelf ROS packages at the lower level and our own inverse kinematics library.

On both the real and the simulated platforms, we use, in addition to other data as detailed in \cref{sec:exp_environments}, RGB images and depth images.
On the real platform, images are captured using a RealSense D435.
The sensor is firmly mounted to the mobile base (see \cref{fig:real_experimental_setup}), ensuring that, when the base moves, the sensor moves with it (see \cref{fig:base_pose_frames_all}).

\begin{figure}[b!]
	\vspace{-3mm}
	\centering
	\includegraphics[width=1.0\columnwidth]{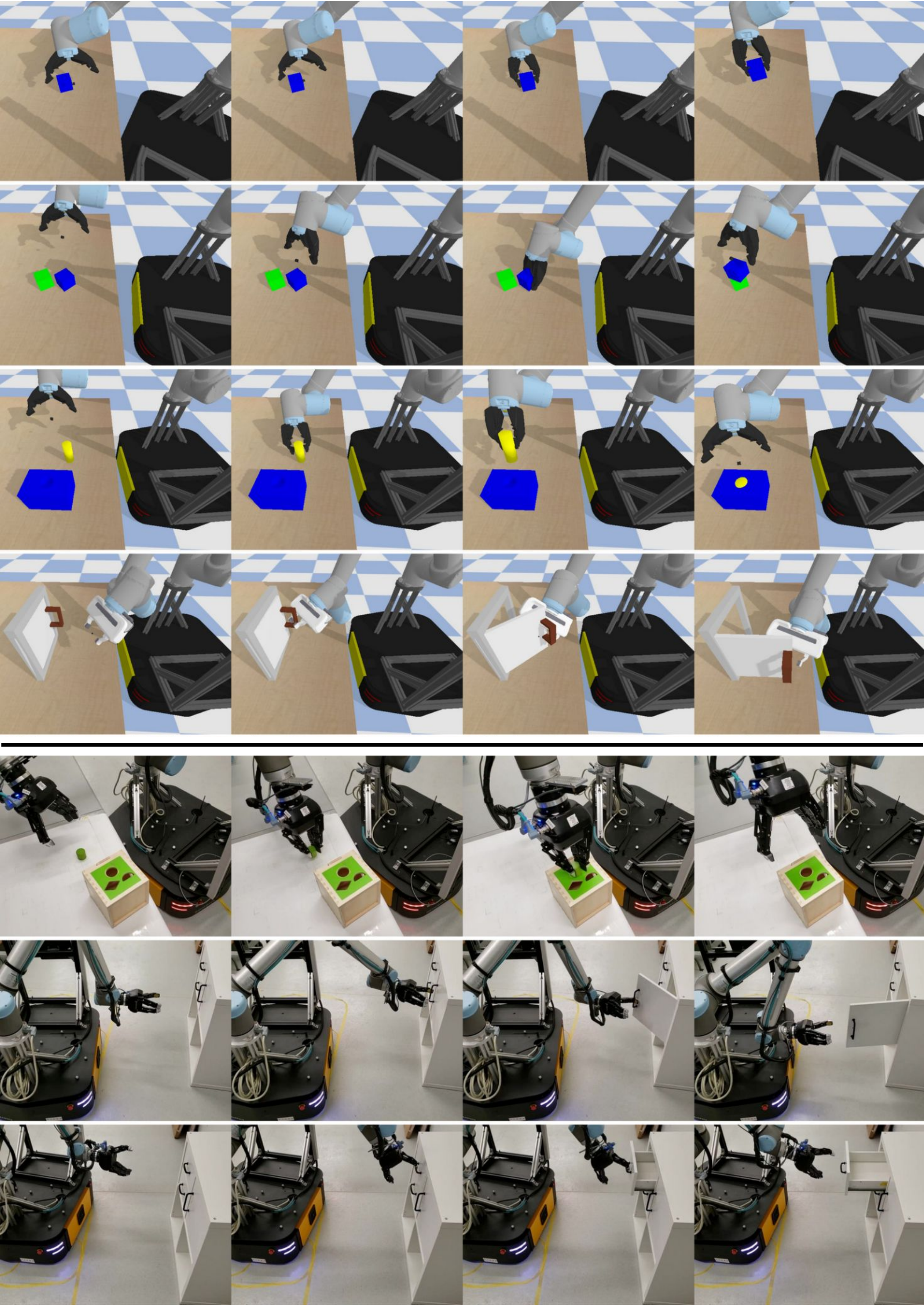}
	\caption{Successful trajectories for tasks studied in this work. From top to bottom: \textit{LiftSim}, \textit{StackSim}, \textit{PickAndInsertSim}, \textit{DoorSim}, \textit{PickAndInsertReal}, \textit{DoorReal}, and \textit{DrawerReal}. See attached video for full examples.}
	\label{fig:example_runs}
\end{figure}

\begin{figure*}[t!]
	\centering
	\smallskip
	\setlength{\fboxsep}{0pt}%
	\setlength{\fboxrule}{1pt}%
	\includegraphics[width=\textwidth]{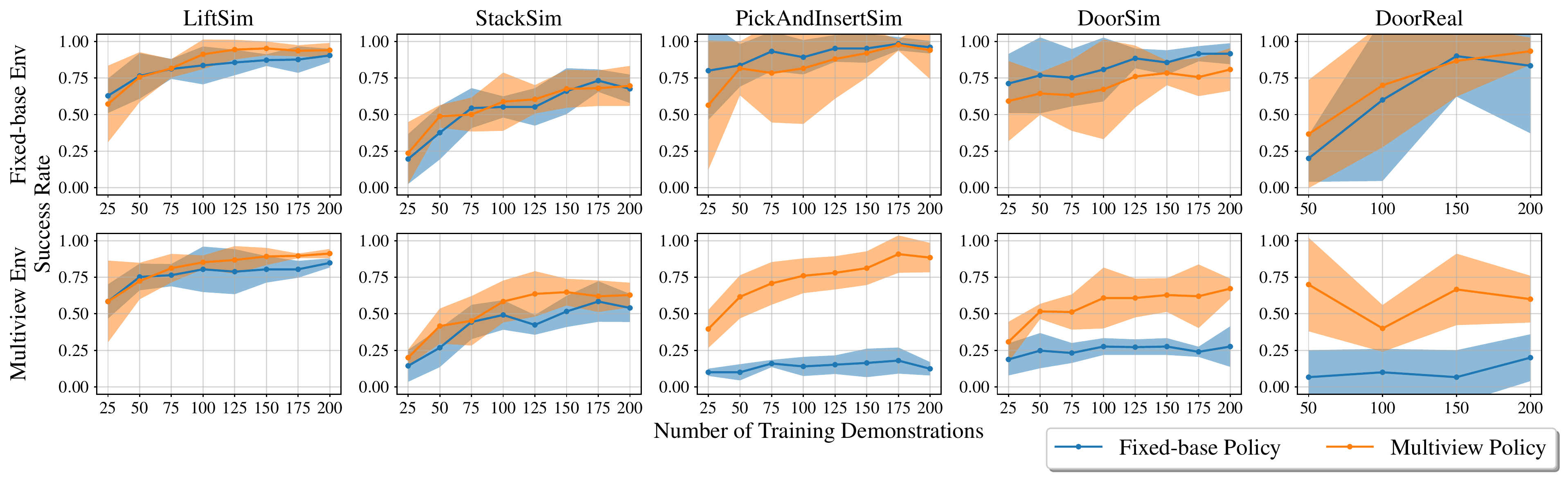}
	\vspace{-6mm}
	\caption{Performance of our policies in environments in which we compared fixed-base with multiview policies in fixed-base (top) and multiview (bottom) environments. The shaded region shows the two-sigma bounds across five policy seeds in simulation, and three in the \textit{DoorReal} environment. The multiview policies, as expected, outperform fixed-base policies in multiview settings, often substantially so, with either no or only minor detriment compared with a fixed-base policy in a fixed-base environment.}
	\label{fig:success_plots}
	\vspace{-3mm}
\end{figure*}

\begin{figure}[b!]
	\vspace{-3mm}
	\centering
	\setlength{\fboxsep}{0pt}%
	\setlength{\fboxrule}{1pt}%
	\includegraphics[width=\columnwidth]{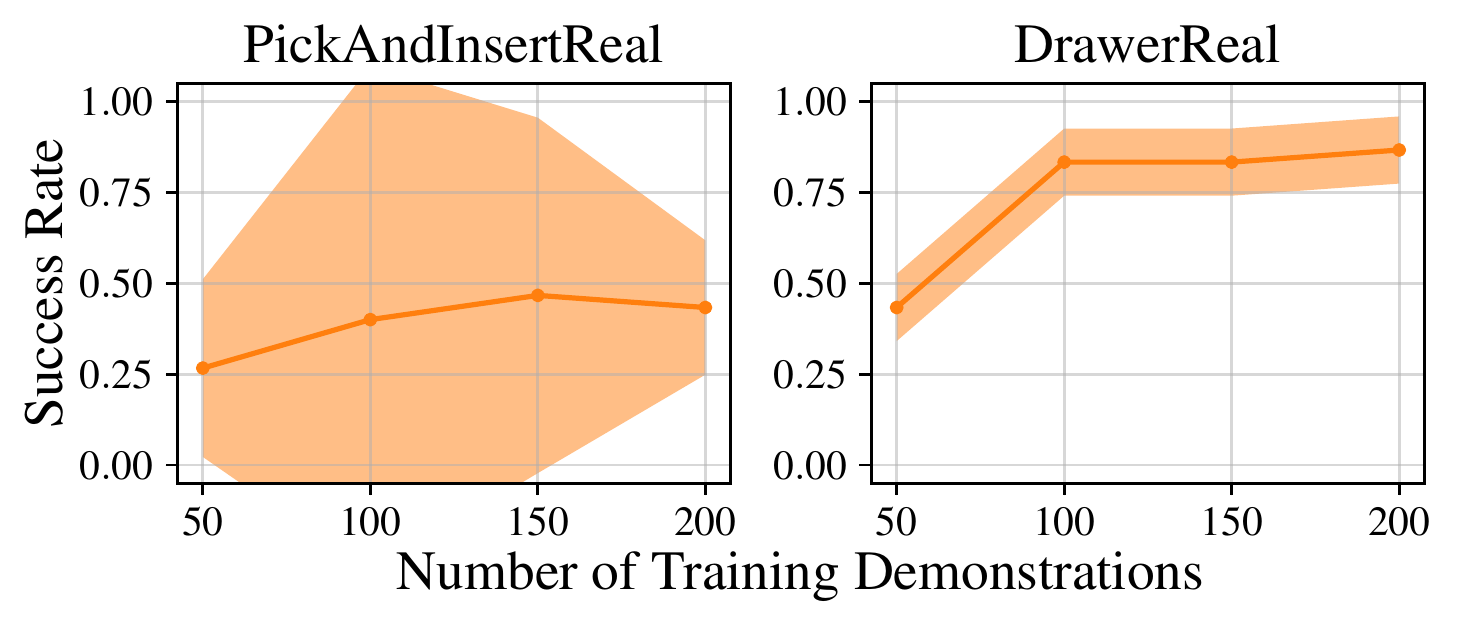}
	\caption{Performance results for the other real environments in which we only tested multiview policies. The shaded region shows the two-sigma bounds across three policy seeds.}
	\label{fig:success_plots_mult_only}
\end{figure}

To collect human demonstrations, we use a single HTC Vive hand controller (see \cref{fig:real_experimental_setup}) and custom-designed software.
For visual feedback, we found that having the user observe the real robot during collection was sufficient.
We also use the force-torque sensor measurements to provide proportional haptic feedback to the demonstrator through the vibration motor in the controller---the vibration amplitude increases with the magnitude of the force and/or torque. 

\subsection{Environments} 
\label{sec:exp_environments}

For a summary of our environments/tasks, see \cref{tab:env_details}. 
Representative images from successful trajectories in each environment are shown in \cref{fig:example_runs}.
All of the tasks' input data include, in addition to RGB and depth images, the current pose of the end-effector in the frame of the robot, provided by forward kinematics and represented as a seven-tuple with position and a unit quaternion for orientation.
As well, each environment that requires actuating the gripper includes the current and previous two positions of each gripper finger.
Finally, the real environments also include force-torque sensor data.

\subsection{Policy Architecture and Training} \label{sec:exp_pol_train}

Our policy networks are inspired by \cite{zhangDeepImitationLearning2018}. 
Specifically, we use a mutli-layer convolutional neural network (CNN) to process the RGB and depth images, take the spatial soft-argmax \cite{levineEndtoendTrainingDeep2016} of the final CNN layer, and concatenate these points with other numerical state information before pushing them through a set of two fully-connected layers.
Our CNN layers are the same as in \cite{zhangDeepImitationLearning2018}, but we use 512 neurons in each of our fully-connected hidden layers.
Crucially, all inputs are available from raw sensor data and our policy does not have access to any privileged state information, including object poses or the relative base pose.
We reduce the resolution of our RGB and depth images to 64$\times$48$\times$3 and 64$\times$48, respectively, and initialize the weights of the RGB layer with weights from ResNet \cite{heDeepResidualLearning2016}.
Empirically, we found greatly reduced variance between the performance of differently-seeded policies by training our policies as ensembles \cite{breimanBaggingPredictors1996}, so each policy is a five-member ensemble, with the final output being the mean output. 
Each member policy is trained with the same data shuffled differently, initialized with different random orthogonal weights (apart from the pretrained weights).

We train the policies using Tensorflow \cite{abadiTensorFlowLargescaleMachine2015} and the Adam optimizer \cite{kingmaAdamMethodStochastic2015} with early stopping, ending training when the validation error on a 20\% holdout set has not improved for 30 epochs. We use a learning rate of 0.001, a mini-batch size of 64, and a maximum of 200 epochs. Our loss function is the mean squared error (\cref{eq:bc_mse}) between the expert and policy action. We did not do a hyperparameter search because our results given these parameters sufficiently answered the questions posed in \cref{sec:introduction}, but presumably, a search could have marginally improved our success rates.

The \textit{LiftSim} environment policies are trained using demonstrations generated from a policy learned with Soft-Actor Critic \cite{haarnojaSoftActorCriticOffPolicy2018} to encourage repeatable experiments. 
Given the high cost of generating autonomous policies for the other tasks, all other tasks use exclusively human-generated data.

\section{Experimental Results} 
\label{sec:experiments}

Our goal in this work is to investigate the performance of multiview policies relative to fixed-view policies on a series of contact-rich manipulation tasks. To do so, we compare the performance of $\pi_m$ and $\pi_f$ in both $\mathcal{T}_m$ and $\mathcal{T}_f$ and on out-of-distribution data. 
To attempt to explain performance gaps, we additionally examine the spatial consistency of the visual features learned by $\pi_m$ and $\pi_f$.

For each task, we collected 200 demonstrations and trained five policies, with different seeds and at multiple demonstration quantities, and finally ran a series of test episodes with held-out initial conditions to evaluate the success rate of each policy.
We trained our policies with increments of 25 demonstrations per policy in simulation, and 50 per policy on the real robot.
We tested our policies with 50 evaluation episodes per policy in simulation, and 10 evaluation episodes per policy on the real robot.

\subsection{Multiview versus Fixed-base}
\label{sec:perf_experiments}

For \textit{LiftSim}, \textit{StackSim}, \textit{PickAndInsertSim}, \textit{DoorSim}, and \textit{DoorReal}, we collected both multiview and fixed-base data on each task and compared performance under four conditions: $\pi_m$ in $\mathcal{T}_m$, $\pi_m$ in $\mathcal{T}_f$, $\pi_f$ in $\mathcal{T}_m$, and $\pi_f$ in $\mathcal{T}_f$ (see \cref{fig:success_plots}).
Notably, as we predicted in \cref{sec:correlation_views}, in the Lift and Stack environments, the multiview policy only provides a marginal benefit over a fixed-base policy in a multiview environment.
For these two environments, $\pi_m$ does not perform any worse than $\pi_f$ in $\mathcal{T}_f$, indicating that a multiview policy can improve performance, and does not appear to cause any detriment.

The benefits of a multiview policy are much clearer in the \textit{PickAndInsertSim}, \textit{DoorSim}, and \textit{DoorReal} environments, where the fixed-base policy fails often in the multiview case, while the multiview policy, as expected, increases in performance with the number of demonstrations. 
Compared with a fixed-base policy, the multiview policy does lose a small amount of performance in the $\mathcal{T}_f$ \textit{DoorSim} task. %
Notably, we do not see the same effect in the real world version of the Door task. 
We suspect that in the real world, it is quite difficult to ensure that the base is in the \textit{exact} same pose it was in during data collection.
Of course, this small deviation is not an issue for the multiview policy and further motivates its use over a fixed-view policy.

\begin{figure*}[t!]
	\centering
	\smallskip
	\setlength{\fboxsep}{0pt}%
	\setlength{\fboxrule}{1pt}%
	\includegraphics[width=\textwidth]{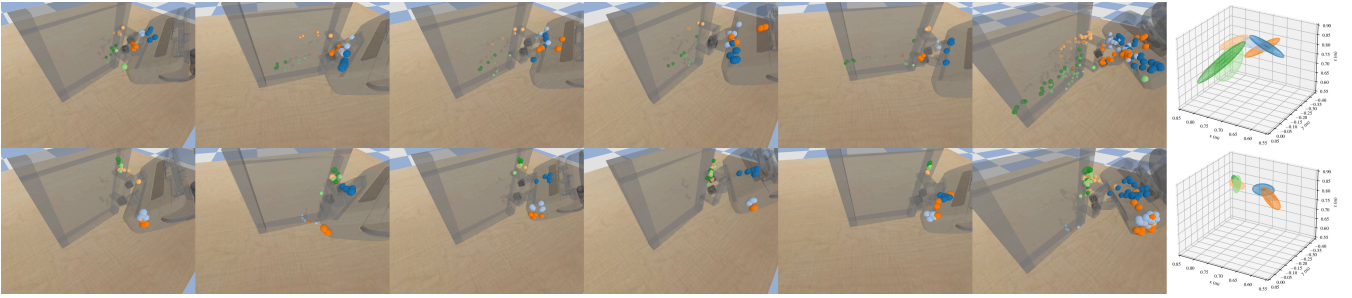}
	\caption{A comparison of the three SSAM points with highest activation on the gripper and the door, reprojected from five different time steps for five different episodes with different viewpoints. The first five images of each row show the locations of the six features for each different viewpoint and the sixth image shows all SSAM points reprojected to a single view. The features in the top diagrams are from $\pi_f$, and display far less spatial consistency than the features in the bottom diagrams from $\pi_m$. The ellipsoids on the right show the two-sigma bounds of the covariance of all of the object-consistent SSAM positions. Each SSAM output corresponds to a different colour. See attached video for more details.}
	\label{fig:feature_analysis}
	\vspace{-3mm}
\end{figure*}

\begin{figure}[b!]
	\vspace{-3mm}
	\centering
	\smallskip
	\setlength{\fboxsep}{0pt}%
	\setlength{\fboxrule}{1pt}%
	\includegraphics[width=\columnwidth]{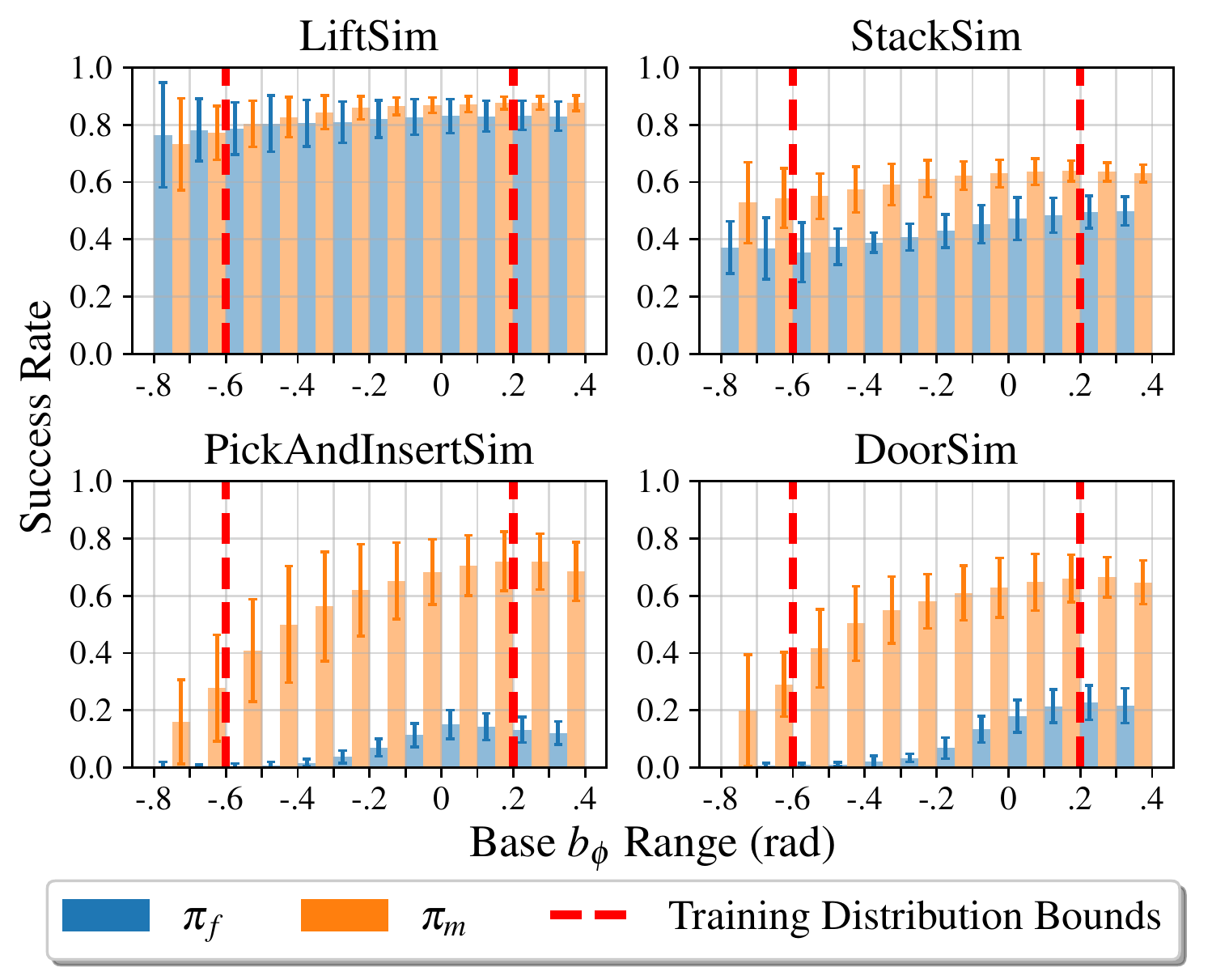}
	\caption{Results of testing multiview policies ($\pi_m$) and fixed-base policies ($\pi_f$), each trained with 200 demonstrations, in multiview environments at a range of angles. The whiskers show the two-sigma bounds across five policy seeds. Here, $\pi_m$ outperforms $\pi_f$, and also shows some ability to perform adequately outside of the training distribution bounds.}
	\label{fig:ood_comparisons}
\end{figure}

The performance results of multiview policies for \textit{PickAndInsertReal} and \textit{DrawerReal} are shown in \cref{fig:success_plots_mult_only}. As is the case for \textit{PickAndInsertSim}, \textit{DoorSim}, and \textit{DoorReal}, we would reasonably expect that a fixed-base policy would not be able to complete these tasks successfully given differing views. 
It is worth noting that the performance variation between differently-seeded policies for \textit{PickAndInsertReal} is relatively high.
We suspect this is due to the difficulty of learning this task in a purely supervised  framework. 
Empirically, many of the failures in this environment were ``near-misses.''

\subsection{Out-of-Distribution (OOD) Experiments} \label{sec:ood_experiments}

To investigate performance on OOD data, we compared the success rate of multiview policies with fixed-view policies given specific base angles, $b_\phi$, as described in \cref{sec:auto_gen_new_views}. 
Our multiview policies were trained in our simulated environments with $b_{\phi} \sim U(-0.6, 0.2)$ (radians), and our fixed-view policies were trained with $b_{\phi} = 0$.
We tested both policies with 12 sets of initial conditions for $b_\phi$: $b_{\phi, \text{range}} = \{ \left[ -0.8, -0.7 \right], \left[ -0.7, -0.6 \right], \dots, \left[ 0.3, 0.4 \right] \}$.
We drew 50 random values from each $b_{\phi, \text{range}}$, and recorded the success rate on these episodes for five seeds of multiview policies and fixed-base policies, each trained with 200 expert demonstrations.
The results are shown in \cref{fig:ood_comparisons}.

As predicted, both types of policies tend to perform roughly equally in each range of angles in \textit{LiftSim}, and, as shown in \cref{sec:perf_experiments}, the \textit{StackSim} multiview policy tends to perform better in general.
For \textit{PickAndInsertSim} and \textit{DoorSim}, a clearer picture emerges to explain the performance difference shown in \cref{fig:success_plots}: the fixed-base policies were trained exclusively at $b_{\phi} = 0$ with no variation in $b_x$ or $b_y$, so $\pi_f$ performance, with even small variations in $b_{\phi}$, falls dramatically compared with the performance of $\pi_f$ on $\mathcal{T}_f$.
The performance of $\pi_f$ continues to deteriorate as $|b_{\phi}|$ increases, while the multiview policies do well throughout the training distribution, with a noticeable negative skew in performance towards the negative angles. 
This reduced performance may occur because our camera is already at an angle to the left of the scene (see \cref{fig:base_pose_frames_all}), so moving it further to the left makes the task particularly challenging.
The multiview policies show some degree of ability to generalize beyond their training distribution, indicating that the multiview policies learn about the geometric relationship between the arm and the objects in the scene.

\subsection{Learned Feature Analysis} 
\label{sec:feature_analysis}

Expanding on our analysis in \cref{sec:correlation_views}, we compare the visual features learned by policies $\pi_f$ and $\pi_m$.
The vision portion of our network terminates with a set of 32 spatial soft-argmax (SSAM) outputs per ensemble member, generated from each of the last convolutional filters, which can be interpreted as points in image space (we refer the reader to \cite{levineEndtoendTrainingDeep2016} for a more detailed explanation of SSAM).
In this section, we refer to SSAM outputs/points interchangeably as features, but unlike traditional features in computer vision (i.e., those used for feature matching), they do not specifically encode a descriptor that can consistently identify exactly the same parts of different images.

The use of SSAM outputs allows us to interpret where the network directs its attention. 
We can therefore observe whether individual SSAM outputs are spatially consistent, which would imply that a view-independent geometric representation has been learned and potentially (partially) explain the generalization capability.
As noted in \cref{sec:correlation_views}, the reuse of information between views would lead to spatially consistent (correlated) features.

For this analysis, we used five random episodes (i.e., with five different views) of each of $\pi_m$ and $\pi_f$ acting in $\mathcal{T}_m$ in our \textit{DoorSim} environment.
In each episode, starting from one time step before the policy initially closed the gripper (attempting to grasp the door handle---arguably the most challenging part of the task), we recorded all of the SSAM points and activation magnitudes from each policy for five time steps.
These SSAM points were then projected into Cartesian space using the known camera intrinsic parameters, depth image data, and the world gripper and door poses. %
Given five episodes and five time steps per episode, each SSAM output produces 25 3D feature points in the world frame.
We sorted the SSAM points by their activation magnitudes, and took the three SSAM points with the highest average activation magnitudes that also showed up at least 20 out of 25 times on either the door or the gripper (as determined using ground-truth information from PyBullet), yielding six representative SSAM outputs in total (three for the door, three for the gripper) for each of $\pi_m$ and $\pi_f$.
Each of these six SSAM features had between 20 and 25 positions on either the gripper or the door. 
We plotted the reprojected locations of these features in \cref{fig:feature_analysis} (with features from $\pi_f$ on the top row and from $\pi_m$ on the bottom row).

The features learned by $\pi_m$ clearly show a smaller degree of spread, and higher spatial correlation, than the features from $\pi_f$.
The spatial correlation of the SSAM layer activations indicates that a degree of view-invariance has been learned without the need to explicitly train using a view-invariance loss or architecture---the policy has learned a visual representation of the task-relevant objects (including the arm) in terms of features that are robust to viewpoint changes.
Our interpretation that $\pi_m$ has learned a degree of true view-invariance is also supported by our results in \cref{sec:ood_experiments}: $\pi_m$ generalizes to viewpoint shifts \textit{beyond} its training distribution.

\section{Conclusion and Future Work}

In this paper, we generated end-to-end policies for challenging, contact-rich tasks involving multiple views.
We demonstrated the benefits of multiview policies through extensive experiments on a mobile manipulation platform in both simulation and in the real world.
Specifically, given the same amount of training data, a multiview policy can be learned with very little, if any, detriment to performance compared with a fixed-base policy and a corresponding fixed-base task. 
Since multiview policies are considerably more flexible than their fixed-based counterparts, we assert that multiview data is potentially always desirable.
As future work, we would like to further investigate methods for reducing the data required to learn effective policies through the use of traditional and learning-based view synthesis techniques or multiview representation learning.

\section*{Acknowledgements}
We gratefully acknowledge the contribution of NVIDIA Corporation, who provided the Titan X GPU used for this research through their Hardware Grant Program.

\bibliographystyle{IEEEcaps}  %
\bibliography{refs_abbr_new_fixed}

\begin{thebibliography}{10}
\def\url#1{}
\csname url@rmstyle\endcsname
\providecommand{\newblock}{\relax}
\providecommand{\bibinfo}[2]{#2}
\providecommand\BIBentrySTDinterwordspacing{\spaceskip=0pt\relax}
\providecommand\BIBentryALTinterwordstretchfactor{4}
\providecommand\BIBentryALTinterwordspacing{\spaceskip=\fontdimen2\font plus
\BIBentryALTinterwordstretchfactor\fontdimen3\font minus
  \fontdimen4\font\relax}
\providecommand\BIBforeignlanguage[2]{{%
\expandafter\ifx\csname l@#1\endcsname\relax
\typeout{** WARNING: IEEEtran.bst: No hyphenation pattern has been}%
\typeout{** loaded for the language `#1'. Using the pattern for}%
\typeout{** the default language instead.}%
\else
\language=\csname l@#1\endcsname
\fi
#2}}

\bibitem{pomerleauALVINNAutonomousLand1989}
D.~A. Pomerleau, ``{{ALVINN}}: {{An Autonomous Land Vehicle}} in a {{Neural
  Network}},'' in \emph{Proc. Ann. Conf. Neural Information Processing Systems
  ({{NIPS}}'89)}, D.~S. Touretzky, Ed., 1989, pp. 305--313.

\bibitem{bojarskiEndEndLearning2016}
M.~Bojarski, \emph{et~al.}, ``\BIBforeignlanguage{en}{End to {{End Learning}}
  for {{Self}}-{{Driving Cars}}},''
  \emph{\BIBforeignlanguage{en}{arXiv:1604.07316 [cs]}}, Apr. 2016.

\bibitem{levineEndtoendTrainingDeep2016}
S.~Levine, C.~Finn, T.~Darrell, and P.~Abbeel, ``End-to-End Training of Deep
  Visuomotor Policies,'' \emph{J. Machine Learning Research}, vol.~17, no.~39,
  pp. 1--40, 2016.

\bibitem{zhangDeepImitationLearning2018}
T.~Zhang, \emph{et~al.}, ``Deep {{Imitation Learning}} for {{Complex
  Manipulation Tasks}} from {{Virtual Reality Teleoperation}},'' in \emph{Proc.
  {{IEEE}} Int. Conf. Robotics and Automation ({{ICRA}}'18)}, May 2018, pp.
  5628--5635.

\bibitem{laskeyDARTNoiseInjection2017}
M.~Laskey, J.~Lee, R.~Fox, A.~D. Dragan, and K.~Goldberg, ``{{DART}}: {{Noise
  Injection}} for {{Robust Imitation Learning}},'' in \emph{Proc. 1st Ann.
  Conf. Robot Learning ({{CoRL}}'17)}, Nov. 2017, pp. 143--156.

\bibitem{iriondoPickPlaceOperations2019}
A.~Iriondo, E.~Lazkano, L.~Susperregi, J.~Urain, A.~Fernandez, and J.~Molina,
  ``\BIBforeignlanguage{en}{Pick and {{Place Operations}} in {{Logistics
  Using}} a {{Mobile Manipulator Controlled}} with {{Deep Reinforcement
  Learning}}},'' \emph{\BIBforeignlanguage{en}{Applied Sciences}}, vol.~9,
  no.~2, p. 348, Jan. 2019.

\bibitem{finnGuidedCostLearning2016}
C.~Finn, S.~Levine, and P.~Abbeel, ``Guided Cost Learning: Deep Inverse Optimal
  Control via Policy Optimization,'' in \emph{Proc. 33rd Int. Conf. Machine
  Learning ({{ICML}}'16)}, ser. {{ICML}}'16, June 2016, pp. 49--58.

\bibitem{codevillaEndtoEndDrivingConditional2018}
F.~Codevilla, M.~M{\"u}ller, A.~L{\'o}pez, V.~Koltun, and A.~Dosovitskiy,
  ``End-to-{{End Driving Via Conditional Imitation Learning}},'' in \emph{Proc.
  {{IEEE}} Int. Conf. Robotics and Automation ({{ICRA}}'18)}, May 2018, pp.
  4693--4700.

\bibitem{bajracharyaMobileManipulationSystem2020}
M.~Bajracharya, \emph{et~al.}, ``A {{Mobile Manipulation System}} for
  {{One}}-{{Shot Teaching}} of {{Complex Tasks}} in {{Homes}},'' in \emph{Proc.
  {{IEEE}} Int. Conf. Robotics and Automation ({{ICRA}}'20)}, May 2020, pp.
  11\,039--11\,045.

\bibitem{wangLearningMobileManipulation2020}
C.~Wang, \emph{et~al.}, ``Learning {{Mobile Manipulation}} through {{Deep
  Reinforcement Learning}},'' \emph{Sensors (Basel, Switzerland)}, vol.~20,
  no.~3, Feb. 2020.

\bibitem{welscheholdLearningMobileManipulation2017}
T.~Welschehold, C.~Dornhege, and W.~Burgard, ``Learning {{Mobile Manipulation
  Actions}} from {{Human Demonstrations}},'' in \emph{Proc. {{IEEE}}/{{RSJ}}
  Int. Conf. Intelligent Robots and Systems ({{IROS}}'17)}, Sep. 2017, pp.
  3196--3201.

\bibitem{kindleWholeBodyControlMobile2020}
J.~Kindle, F.~Furrer, T.~Novkovic, J.~J. Chung, R.~Siegwart, and J.~Nieto,
  ``Whole-{{Body Control}} of a {{Mobile Manipulator}} Using {{End}}-to-{{End
  Reinforcement Learning}},'' \emph{arXiv:2003.02637 [cs]}, Feb. 2020.

\bibitem{laskeyLearningRobustBed2017}
M.~Laskey, C.~Powers, R.~Joshi, A.~Poursohi, and K.~Goldberg, ``Learning
  {{Robust Bed Making}} Using {{Deep Imitation Learning}} with {{DART}},''
  \emph{arXiv:1711.02525 [cs]}, Nov. 2017.

\bibitem{aminiLearningRobustControl2020}
A.~Amini, \emph{et~al.}, ``Learning {{Robust Control Policies}} for
  {{End}}-to-{{End Autonomous Driving From Data}}-{{Driven Simulation}},''
  \emph{IEEE Robotics and Automation Letters}, vol.~5, no.~2, pp. 1143--1150,
  Apr. 2020.

\bibitem{eslamiNeuralSceneRepresentation2018}
S.~M.~A. Eslami, \emph{et~al.}, ``\BIBforeignlanguage{en}{Neural Scene
  Representation and Rendering},'' \emph{\BIBforeignlanguage{en}{Science}},
  vol. 360, no. 6394, pp. 1204--1210, June 2018.

\bibitem{sermanetTimeContrastiveNetworksSelfSupervised2018}
P.~Sermanet, \emph{et~al.}, ``Time-{{Contrastive Networks}}:
  {{Self}}-{{Supervised Learning}} from {{Video}},'' in \emph{{{IEEE}} Int.
  Conf. Robotics and Automation ({{ICRA}}'18)}, May 2018, pp. 1134--1141.

\bibitem{dwibediLearningActionableRepresentations2018}
D.~Dwibedi, J.~Tompson, C.~Lynch, and P.~Sermanet, ``Learning {{Actionable
  Representations}} from {{Visual Observations}},'' in \emph{The
  {{IEEE}}/{{RSJ}} Int. Conf. Intelligent Robots and Systems ({{IROS}}'18)},
  Oct. 2018, pp. 1577--1584.

\bibitem{maedaVisualTaskProgress2020}
G.~Maeda, J.~V{\"a}{\"a}t{\"a}inen, and H.~Yoshida, ``Visual {{Task Progress
  Estimation}} with {{Appearance Invariant Embeddings}} for {{Robot Control}}
  and {{Planning}},'' in \emph{The {{IEEE}}/{{RSJ}} Int. Conf. Intelligent
  Robots and Systems ({{IROS}}'20)}, Oct. 2020, pp. 7941--7948.

\bibitem{florenceDenseObjectNets2018}
P.~R. Florence, L.~Manuelli, and R.~Tedrake, ``\BIBforeignlanguage{en}{Dense
  {{Object Nets}}: {{Learning Dense Visual Object Descriptors By}} and {{For
  Robotic Manipulation}}},'' in \emph{\BIBforeignlanguage{en}{The 2nd Ann.
  Conf. Robot Learning ({{CoRL}}'18)}}, Oct. 2018, pp. 373--385.

\bibitem{florenceSelfSupervisedCorrespondenceVisuomotor2020}
P.~Florence, L.~Manuelli, and R.~Tedrake, ``Self-{{Supervised Correspondence}}
  in {{Visuomotor Policy Learning}},'' \emph{IEEE Robotics and Automation
  Letters}, vol.~5, no.~2, pp. 492--499, Apr. 2020.

\bibitem{sadeghiSim2RealViewpointInvariant2018}
F.~Sadeghi, A.~Toshev, E.~Jang, and S.~Levine, ``{{Sim2Real Viewpoint Invariant
  Visual Servoing}} by {{Recurrent Control}},'' in \emph{Proc. {{IEEE}}/{{CVF}}
  Conf. Computer Vision and Pattern Recognition ({{CVPR}}'18)}, June 2018, pp.
  4691--4699.

\bibitem{tobinDomainRandomizationTransferring2017}
J.~Tobin, R.~Fong, A.~Ray, J.~Schneider, W.~Zaremba, and P.~Abbeel, ``Domain
  Randomization for Transferring Deep Neural Networks from Simulation to the
  Real World,'' in \emph{Proc. {{IEEE}}/{{RSJ}} Int. Conf. Intelligent Robots
  and Systems ({{IROS}}'17)}, Sep. 2017, pp. 23--30.

\bibitem{bainFrameworkBehaviouralCloning1996}
M.~Bain and C.~Sammut, ``A {{Framework}} for {{Behavioural Cloning}},'' in
  \emph{Machine {{Intelligence}} 15}, 1996, pp. 103--129.

\bibitem{rossReductionImitationLearning2011}
S.~Ross, G.~J. Gordon, and D.~Bagnell, ``A {{Reduction}} of {{Imitation
  Learning}} and {{Structured Prediction}} to {{No}}-{{Regret Online
  Learning}},'' in \emph{Proc. 14th Int. Conf. Artificial Intelligence and
  Statistics ({{AISTATS}}'11)}, 2011, pp. 627--635.

\bibitem{ablettFightingFailuresFIRE2020}
T.~Ablett, F.~Mari{\'c}, and J.~Kelly, ``Fighting {{Failures}} with {{FIRE}}:
  {{Failure Identification}} to {{Reduce Expert Burden}} in
  {{Intervention}}-{{Based Learning}},'' \emph{arXiv:2007.00245 [cs]}, Aug.
  2020.

\bibitem{bellmanDynamicProgramming1957}
R.~Bellman, \emph{Dynamic Programming}, ser. Dover Books on Computer Science
  Series, 1957.

\bibitem{coumans2019}
E.~Coumans and Y.~Bai, ``{{PyBullet}}, a {{Python}} Module for Physics
  Simulation for Games, Robotics and Machine Learning,'' 2016.

\bibitem{heDeepResidualLearning2016}
K.~He, X.~Zhang, S.~Ren, and J.~Sun, ``Deep {{Residual Learning}} for {{Image
  Recognition}},'' in \emph{Proc. {{IEEE}} Conf. Computer Vision and Pattern
  Recognition ({{CVPR}}'16)}, June 2016, pp. 770--778.

\bibitem{breimanBaggingPredictors1996}
L.~Breiman, ``\BIBforeignlanguage{en}{Bagging {{Predictors}}},''
  \emph{\BIBforeignlanguage{en}{Machine Learning}}, vol.~24, no.~2, pp.
  123--140, Aug. 1996.

\bibitem{abadiTensorFlowLargescaleMachine2015}
M.~Abadi, \emph{et~al.}, ``{{TensorFlow}}: {{Large}}-Scale Machine Learning on
  Heterogeneous Systems,'' 2015.

\bibitem{kingmaAdamMethodStochastic2015}
D.~P. Kingma and J.~Ba, ``Adam: {{A Method}} for {{Stochastic Optimization}},''
  in \emph{Proc. Int. Conf. Learning Representations ({{ICLR}}'15)}, May 2015.

\bibitem{haarnojaSoftActorCriticOffPolicy2018}
T.~Haarnoja, A.~Zhou, P.~Abbeel, and S.~Levine, ``\BIBforeignlanguage{en}{Soft
  {{Actor}}-{{Critic}}: {{Off}}-{{Policy Maximum Entropy Deep Reinforcement
  Learning}} with a {{Stochastic Actor}}},'' in
  \emph{\BIBforeignlanguage{en}{Proc. 35th Int. Conf. Machine Learning
  ({{ICML}}'18)}}, July 2018, pp. 1861--1870.

\end{thebibliography}

\end{document}